\documentclass[runningheads]{llncs}
\usepackage[T1]{fontenc}
\usepackage{graphicx}
\usepackage{url}
\begin{document}
\title{Unsupervised Method for Intra-patient Registration of Brain Magnetic Resonance Images based on Objective Function Weighting by Inverse Consistency: Contribution to the BraTS-Reg Challenge}
\titlerunning{Unsupervised Method for Intra-patient Registration based on Inverse Consistency}

\author{
Marek Wodzinski\inst{1,2}\orcidID{0000-0002-8076-6246}
\and
Artur Jurgas\inst{1,2}\orcidID{0000-0002-9057-9428}
\and
Niccol\`{o} Marini\inst{1,3}\orcidID{0000-0002-5273-5741}
\and
Manfredo Atzori\inst{1,4}\orcidID{0000-0001-5397-2063}
\and
Henning M\"{u}ller\inst{1,5}\orcidID{0000-0001-6800-9878}
}
\authorrunning{M. Wodzinski et al.}
%
\institute{University of Applied Sciences Western Switzerland\\ Information Systems Institute, Sierre, Switzerland \\
\and
AGH University of Science and Technology\\ Department of Measurement and Electronics, Krakow, Poland \\
\and
Department of Computer Science, University of Geneva, Geneva, Switzerland \\
\and
Department of Neuroscience, University of Padova, Padova, Italy \\
\and
Medical Faculty, University of Geneva, Geneva, Switzerland \\
\email{wodzinski@agh.edu.pl}\\
\email{marek.wodzinski@hevs.ch}\\
}
\maketitle              
\begin{abstract}
Registration of brain scans with pathologies is difficult, yet important research area. The importance of this task motivated researchers to organize the BraTS-Reg challenge, jointly with IEEE ISBI 2022 and MICCAI 2022 conferences. The organizers introduced the task of aligning pre-operative to follow-up magnetic resonance images of glioma. 
The main difficulties are connected with the missing data leading to large, nonrigid, and noninvertible deformations. In this work, we describe our contributions to both the editions of the BraTS-Reg challenge. The proposed method is based on combined deep learning and instance optimization approaches. 
First, the instance optimization enriches the state-of-the-art LapIRN method to improve the generalizability and fine-details preservation. Second, an additional objective function weighting is introduced, based on the inverse consistency.
The proposed method is fully unsupervised and exhibits high registration quality and robustness. The quantitative results on the external validation set are: (i) IEEE ISBI 2022 edition: 1.85, and 0.86, (ii) MICCAI 2022 edition: 1.71, and 0.86, in terms of the mean of median absolute error and robustness respectively. The method scored the 1st place during the IEEE ISBI 2022 version of the challenge and the 3rd place during the MICCAI 2022. Future work could transfer the inverse consistency-based weighting directly into the deep network training.

\end{abstract}

\section{Introduction}
\label{sec:intro}

Registration of brain scans acquired using magnetic resonance imaging (MRI) is an active research area. Numerous works addressed the challenge by enforcing diffeomorphic deformations, using both classical~\cite{syn}, as well as deep learning-based (DL) approaches~\cite{u_voxemorph_diffeo_2,u2,mok,synthmorph}. However, it is still unclear what is the best approach for registering scans containing pathologies. 

To answer this question, a Brain Tumor Sequence Registration (BraTS-Reg) challenge was organized in conjunction with IEEE ISBI 2022 and MICCAI 2022 conferences~\cite{dataset}. The challenge organizers proposed a task dedicated to registration of pre-operative MRI scans of patients diagnosed with a brain glioma, to follow-up scans of the same patient acquired after the treatment.

The surgical resection introduces the problem of missing data where the structure of interest is absent in one of the registered images. The problem of missing data in the registration of brain scans induces two main challenges: (i) large, nonrigid deformations caused by the tumor resection (depending on its size), and (ii) non-invertibility of the displacement field in the tumor area (the displacement field is unable to create new structures).

\textbf{Contribution:} In this work, we present our contribution to both the editions of the BraTS-Reg challenge~\cite{dataset}. We propose an unsupervised, hybrid approach, based on a deep neural network followed by an instance optimization (IO) attempting to address the challenge connected with large, nonrigid deformations. We introduce an additional, unsupervised inverse consistency-based weighting of the objective function. We present our motivations behind the method design and show the high quality of the registration in terms of the median absolute error (MAE) and robustness. 

\section{Methods}
\label{sec:methods}

\subsection{Overview and preprocessing}
\begin{figure*}[!htb]
	\centering
    \includegraphics[scale=0.6]{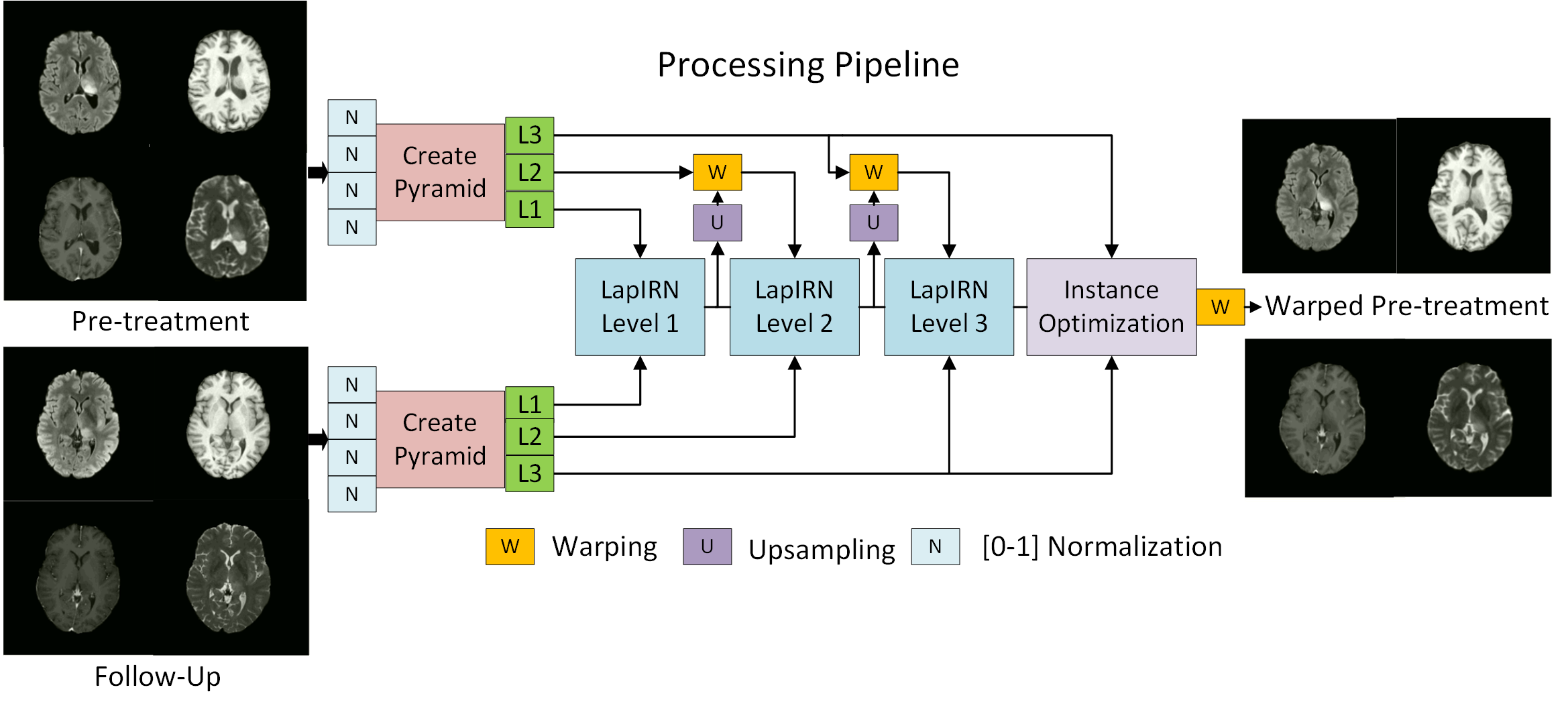}
    \caption{The IEEE ISBI 2022 registration pipeline. Please note that some connections between LapIRN are omitted for the presentation clarity.}   
    \label{fig:pipeline}
\end{figure*}

\begin{figure*}[!htb]
	\centering
    \includegraphics[scale=0.6]{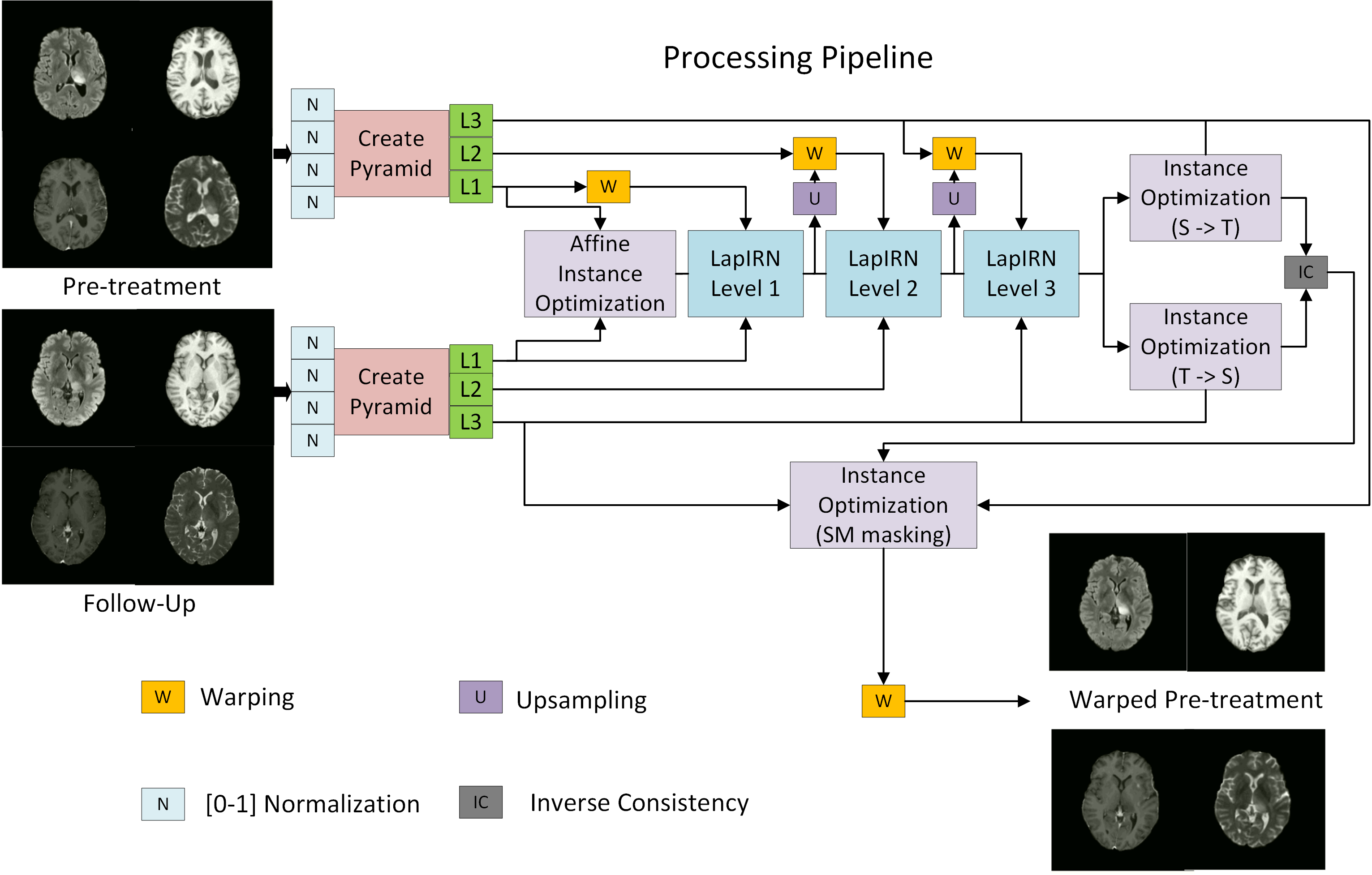}
    \caption{The MICCAI 2022 registration pipeline. Please note that some connections between LapIRN are omitted for the presentation clarity. Note that during the 2nd edition of the challenge the pipeline starts with an initial affine registration.}   
    \label{fig:pipeline_miccai}
\end{figure*}

The proposed method consists of the following steps: (i) an initial, instance optimization-based affine registration, (ii) calculating the initial displacement field using a modified version of the LapIRN~\cite{mok2}, (iii) tuning the displacement field by the instance optimization (IO) based, nonrigid registration method~\cite{btb}, and (iv - MICCAI edition only) an objective function weighting based on the inverse consistency. The ISBI 2022 pipeline is shown in Figure~\ref{fig:pipeline} and the MICCAI 2022 pipeline is shown in Figure~\ref{fig:pipeline_miccai}.

The basic preprocessing was performed offline by challenge organizers and consisted of resampling the images to the same resolution. The processing pipeline starts with creating the source and target tensors. The tensors are created by concatenating all available modalities in the channel dimension, for both the source and target images. This results in two 4x240x240x155 volumes. The volumes are normalized channel-wise to [0-1] intensity values. We use the pre-operative scan as the source volume and the follow-up scan as the target volume. The challenge design requires to warp the landmarks annotated in follow-up scans and we decided to not attempt to invert the noninvertible displacement fields, neither to voxelize the landmarks.

\subsection{Affine Registration}

The process starts with an initial instance optimization-based (IO), iterative affine registration. Even though the volumes were group-wise aligned to the same space by the challenge organizers during preprocessing, the pair-wise affine registration was still beneficial to improve the initial alignment quality.

The affine registration is performed by iterative, IO method with the local normalized cross correlation as the objective function and the Adam as the optimizer. The method directly optimizes the 3-D affine transformation matrix. The proposed method is multilevel, however, in practice it is enough to perform the affine registration at the coarsest resolution level.

\subsection{Deep Network - LapIRN}

The preprocessed volumes are directly passed to a slightly modified LapIRN~\cite{mok2}. The LapIRN is a multi-level network based on resolution pyramids to enlarge the receptive field. We decided to use the LapIRN because it won two editions of the Learn2Reg challenge by a huge margin and confirmed its ability to recover large, nonrigid deformations~\cite{hering2021learn2reg}.

We use the LapIRN to optimize the following objective function:
\begin{equation}
    O_{REG}(S, T, u) = \sum_{i=1}^{N} \frac{1}{2^{(N-i)}}(NCC(S_i \circ u_i, T_i) + \theta_i Reg(u_i)),
\end{equation}
where $S_i, T_i$ are the pre-operative and follow-up scans at i-th resolution level respectively, $u_i$ is the calculated displacement field, $\theta_i$ denotes the regularization coefficient at i-th level, $NCC$ denotes the channel-wise normalized cross-correlation, $Reg$ is the diffusive regularization, $\circ$ denotes the warping operation, and $N$ is the number of pyramid levels. 

We have reimplemented the network to have full control over the training process. We added group normalization due to the small batch size and deleted the scaling and squaring layers because we did not want to force diffeomorphic properties, nor to oversmooth the displacement fields. Even though it would be reasonable to calculate diffeomorphic displacements outside the tumor area, the smoothness of the deformation field was not taken into account to rank the submissions. 

\subsection{Instance Optimization}

The LapIRN alone does not provide satisfactory results. Therefore, the displacement field calculated by LapIRN is passed as an initial transformation to a multi-level nonrigid instance optimization (IO) module. The module fine-tunes the displacement field by an iterative optimization implemented using PyTorch autograd. In the original formulation, the method optimizes the same cost function as the LapIRN.

We decided to use the IO because: (i) it improves the ability to register fine details since the displacement fields are optimized for each case separately, (ii) only the unsupervised objective function is used during LapIRN training, and (iii) the registration time is not scored during the challenge. The IO increases the registration time by several seconds which is usually acceptable in applications not requiring real-time alignment.

\subsection{Inverse Consistency-based Weighting}
To keep into account the missing data, we introduce an objective function weighting based on the inverse consistency. After the initial registration pipeline, the source and target images are bidirectionally registered again. The registration is performed by several iterations of the IO. Then, the displacement fields are composed and the inverse consistency is calculated. The inverse consistency error is defined as:
\begin{equation}
      IC_{err}(u_{st}, u_{ts}) = u_{st}(i) \circ u_{ts}(i),  
\end{equation}
where $u_{st}, u_{ts}$ are the displacement fields from source to target and from target to source respectively, and the $\circ$ denotes the displacement field composition.

\begin{figure*}[!htb]
	\centering
    \includegraphics[scale=0.7]{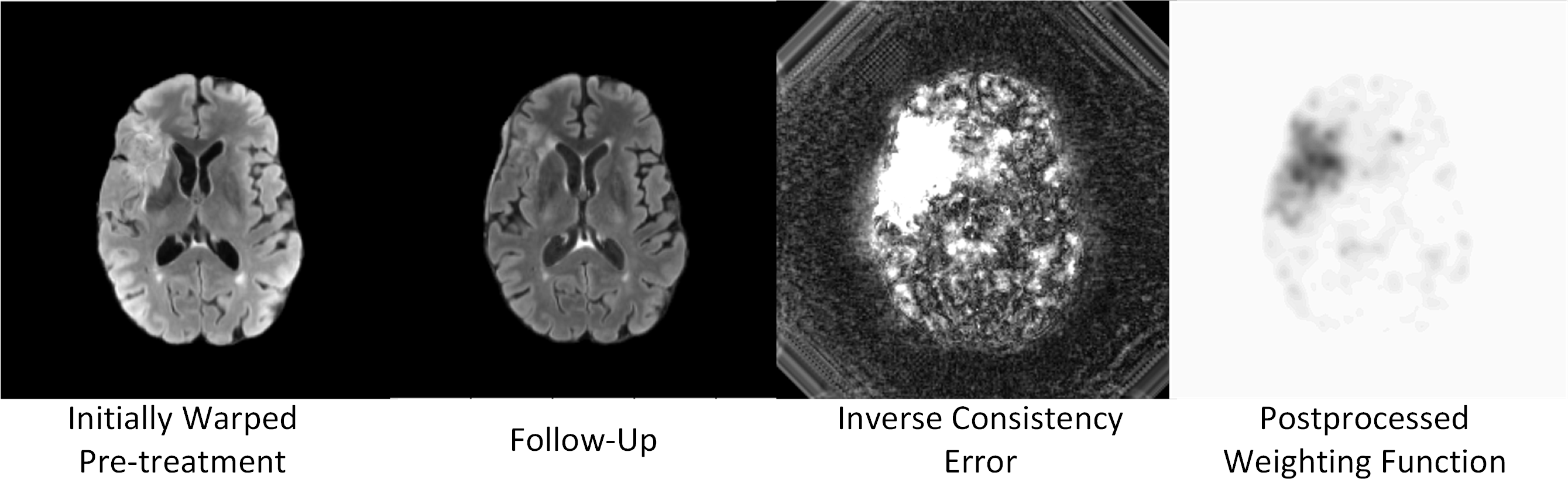}
    \caption{Visualization of the objective function weighting. Note that small values of the weighting function are close to the tumor.}   
    \label{fig:IC_VIS}
\end{figure*}

The inverse consistency is normalized to [0-1] range, raised to a given power, negated and smoothed using Gaussian filter with a predefined sigma. Both the sigma and the power are hyperparameters of the method. An exemplary visualization of the exemplary weighting mask is shown in Figure~\ref{fig:IC_VIS}. Importantly, the weighting mask is applied both to the similarity measure and the regularization term. We decided to weight the regularization term because the displacement field is supposed to model tumor resection by the volume contraction which is prohibited by large values of the regularization coefficient.

The intuition behind this approach is connected with the fact that regions without direct correspondences should have large values of the inverse consistency error. Thus, the information can be used to unsupervisedly detect non-corresponding regions and utilize this information in the objective function weighting.

\subsection{Dataset and Experimental Setup}

The multi-institutional dataset was curated and pre-processed retrospectively by the challenge organizers~\cite{dataset}. The dataset consists of pre-operative and follow-up pairs (of the same patient) diagnosed and treated for glioma. There are four MRI sequences for each pair: (i) T1-weighted (T1), (ii) contrast-enhanced T1-weighted (T1-CE), (iii) T2-weighted (T2), and (iv) fluid-attenuated inversion recovery (FLAIR).

The training set consists of 140 pairs annotated with landmarks varying from 6 to 50 per pair. The landmarks are defined using anatomical markers e.g. midline of the brain, blood vessel bifurcations, and anatomical shape of the cortex. We separated 5\% of the training set as an internal validation set not used during the network training.

The external validation set consists of 20 pairs, following the structure of the training set, however, without releasing the landmarks for the pre-operative scans. The evaluation on the external validation set is possible using the challenge infrastructure only. There is also an external test set, however, unavailable for the challenge participants. The test set results will be published in the follow-up challenge summary publication based on container submissions. Therefore, in this paper, we report results for the external validation set only.

All experiments are implemented in PyTorch, extended by PyTorch Lightning~\cite{pylightning}. The LapIRN is trained using three pyramid levels. First, the first level is trained until convergence (with frozen weights of the remaining levels), followed by unfreezing the second level, and then finally the third level. We pre-train the network by registering the pairs in a cross-case manner, resulting in almost 18,000 training pairs. The network is then fine-tuned using the pairs representing scans of the same patient. Each iteration consists of registering 100 pairs with a batch size equal to 1. The weighted sum of local channel-wise NCC and diffusive regularization are used as the objective function. The pretraining is performed with a window size equal to 7 pixels, then changed to 3 pixels during the fine-tuning. The regularization coefficients are equal to 1000, 2000, and 4000 in the 1st, 2nd and 3rd pyramid level respectively. The initial learning rate is 0.003 and decays by a factor of 0.99 per iteration. No information about the anatomical landmarks is used during training.

The instance optimization, following the DL-based registration, is run using two resolution levels, for 40 and 20 iterations respectively. All the IO runs share the same hyperparameters. The objective function is the same as during training LapIRN and the hyperparameters are as follows: (i) NCC window size is equal to 3 pixels, (ii) the regularization coefficients are equal to 12500 and 25000 respectively, (iii) the initial learning rate is equal to 0.003. During the last instance optimization call, the inverse consistency is used to perform the similarity measure weighting. The sigma and the power of the inverse consistency map postprocessing are both equal to 2. The source code is openly available at~\cite{source_code}.

\section{Results}
\label{sec:results}

The registration is evaluated quantitatively in terms of robustness and median absolute error (MAE) between the anatomical landmarks. The robustness is defined as the ratio of landmarks with improved alignment after registration to all landmarks. The challenge organizers decided to not score the registration time and the deformation field smoothness is used only in case of tie~\cite{dataset}.

\begin{figure*}[!htb]
	\centering
    \includegraphics[scale=1.0]{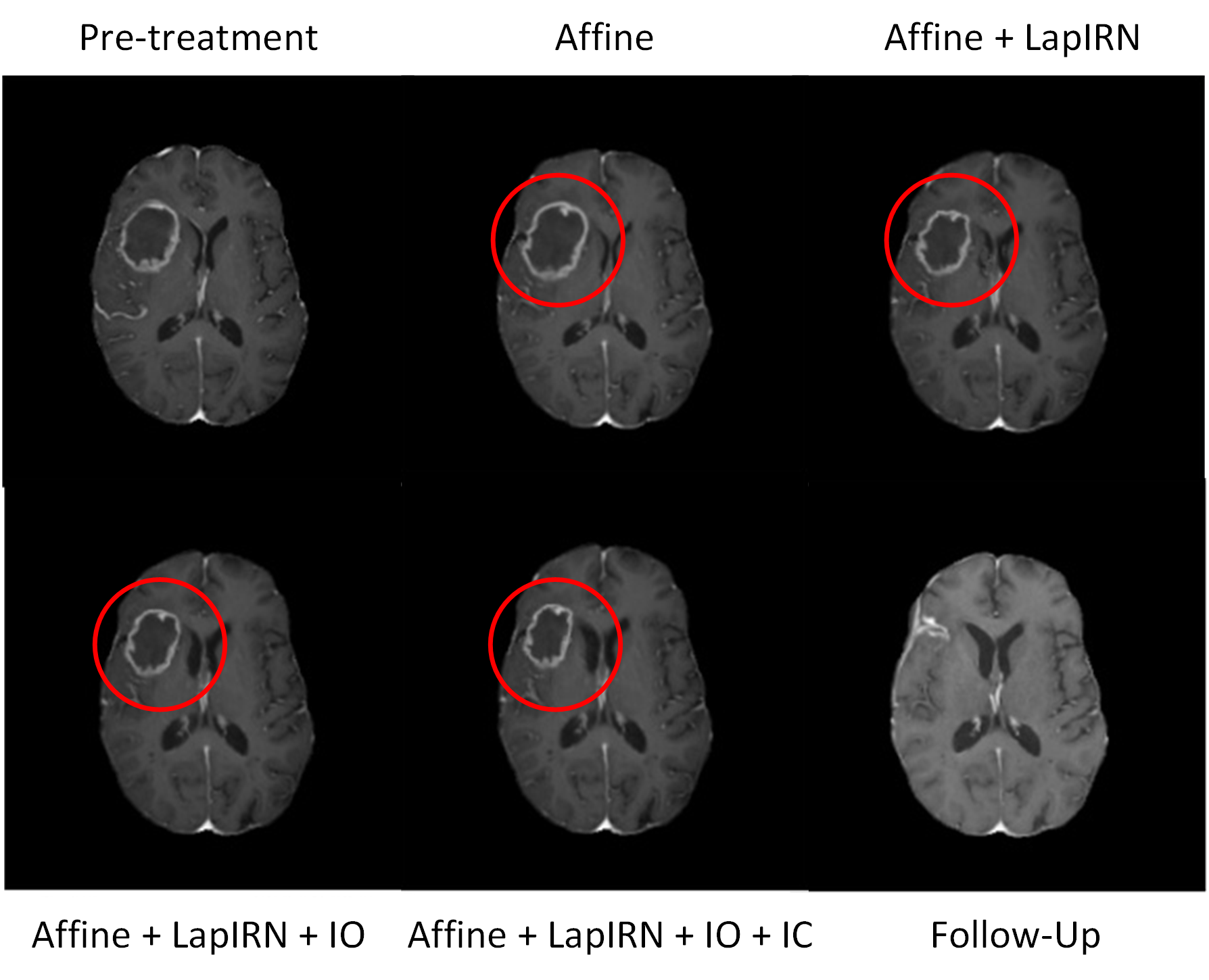}
    \caption{Exemplary visualization of the registration results. Please note that the inverse consistency-based weighting improves the qualitative results close to the tumor volume. The tumor volume decreases and the quality of registration of fine-details around the tumor is improved. This improvement is not captured by the quantitative results based on the target registration error.}   
    \label{fig:vis_results}
\end{figure*}

We show the MAE and robustness statistics on the external validation set in Table~\ref{tab:results} and Table~\ref{tab:results_miccai} for the IEEE ISBI and MICCAI editions respectively. We present the results before the registration, after the affine registration, for the IO and DL-based registration applied separately, for the proposed hybrid approach, and for the hybrid approach extended by unsupervised cost function weighting. These results come from the challenge evaluation platform. An exemplary visualization of the registration results are shown in Figure~\ref{fig:vis_results}.

The proposed method scored the first and the third place during the IEEE ISBI 2022 and MICCAI 2022 editions, respectively (however, without statistically significant differences compared to the remaining methods on the podium). 

The average registration time (RTX GeForce 3090 Ti) on the external validation set is: 1.3, 1.4, 21.8, 57.9 seconds respectively for the Affine, Affine + LapIRN, Affine + LapIRN + IO, Affine + LapIRN + IO + IC version respectively. The registration time is increased by the IO modules that could be further optimized, however, in practice the real-time alignment is not crucial for the discussed registration task.

\begin{table*}[!htb]
\centering
\caption{Quantitative results on the external validation set during the IEEE ISBI 2022 edition (MAE - Median Absolute Error).}
\renewcommand{\arraystretch}{1.0}
\footnotesize
\resizebox{0.99\textwidth}{!}{%
\begin{tabular}{ccccccccc}
\label{tab:results}
Case & \multicolumn{2}{c}{Initial} &
\multicolumn{2}{c}{IO} & \multicolumn{2}{c}{LapIRN} &
\multicolumn{2}{c}{LapIRN + IO}
\tabularnewline
\multicolumn{1}{c}{ID} & \multicolumn{1}{c}{MAE} & \multicolumn{1}{c}{Robustness} & \multicolumn{1}{c}{MAE} & \multicolumn{1}{c}{Robustness} & \multicolumn{1}{c}{MAE} & \multicolumn{1}{c}{Robustness} & \multicolumn{1}{c}{MAE} & \multicolumn{1}{c}{Robustness}
\tabularnewline
\hline
141 & \textit{13.50} &  \textit{-} & \textit{2.06} & \textit{1.00} & \textit{2.78} & \textit{1.00} & \textbf{\textit{1.94}} & \textit{1.00} \tabularnewline
142 & \textit{14.00} &  \textit{-} & \textit{4.13} & \textit{0.86} & \textit{3.14} & \textit{0.86} & \textbf{\textit{2.82}} & \textit{1.00} \tabularnewline
143 & \textit{16.00} &  \textit{-} & \textit{2.00} & \textit{1.00} & \textit{4.51} & \textit{1.00} & \textbf{\textit{1.76}} & \textit{1.00} \tabularnewline
144 & \textit{15.00} &  \textit{-} & \textbf{\textit{3.54}} & \textit{1.00} & \textit{5.20} & \textit{1.00} & \textit{4.12} & \textit{1.00} \tabularnewline
145 & \textit{17.00} &  \textit{-} & \textit{1.54} & \textit{1.00} & \textit{2.96} & \textit{1.00} & \textbf{\textit{1.19}} & \textit{1.00} \tabularnewline
146 & \textit{17.00} &  \textit{-} & \textit{2.26} & \textit{1.00} & \textit{3.56} & \textit{1.00} & \textbf{\textit{2.24}} & \textit{1.00} \tabularnewline
147 & \textit{1.50} &  \textit{-} & \textit{2.08} & \textit{0.40} & \textit{2.41} & \textit{0.50} & \textbf{\textit{1.79}} & \textit{0.45} \tabularnewline
148 & \textit{3.50} &  \textit{-} & \textbf{\textit{1.30}} & \textit{0.90} & \textit{1.99} & \textit{0.85} & \textit{1.63} & \textit{0.80} \tabularnewline
149 & \textit{9.00} &  \textit{-} & \textit{1.66} & \textit{1.00} & \textit{1.91} & \textit{1.00} & \textbf{\textit{1.62}} & \textit{1.00} \tabularnewline
150 & \textit{4.00} &  \textit{-} & \textit{1.66} & \textit{0.63} & \textit{2.34} & \textit{0.68} & \textbf{\textit{1.32}} & \textit{0.63} \tabularnewline
151 & \textit{3.00} &  \textit{-} & \textit{1.40} & \textit{0.80} & \textit{1.39} & \textit{0.70} & \textbf{\textit{1.17}} & \textit{0.80} \tabularnewline
152 & \textit{5.00} &  \textit{-} & \textbf{\textit{1.43}} & \textit{0.95} & \textit{1.68} & \textit{0.95} & \textit{1.46} & \textit{0.95} \tabularnewline
153 & \textit{2.00} &  \textit{-} & \textbf{\textit{1.47}} & \textit{0.67} & \textit{1.61} & \textit{0.67} & \textit{1.96} & \textit{0.67} \tabularnewline
154 & \textit{2.00} &  \textit{-} & \textit{2.09} & \textit{0.55} & \textit{2.12} & \textit{0.45} & \textbf{\textit{2.03}} & \textit{0.55} \tabularnewline
155 & \textit{2.00} &  \textit{-} & \textit{1.93} & \textit{0.47} & \textit{2.34} & \textit{0.42} & \textbf{\textit{1.92}} & \textit{0.53} \tabularnewline
156 & \textit{7.00} &  \textit{-} & \textbf{\textit{1.74}} & \textit{1.00} & \textit{2.14} & \textit{1.00} & \textit{1.80} & \textit{1.00} \tabularnewline
157 & \textit{10.00} &  \textit{-} & \textit{2.20} & \textit{1.00} & \textit{2.17} & \textit{1.00} & \textbf{\textit{1.22}} & \textit{1.00} \tabularnewline
158 & \textit{4.50} &  \textit{-} & \textit{1.43} & \textit{0.90} & \textit{1.81} & \textit{0.90} & \textbf{\textit{1.19}} & \textit{1.00} \tabularnewline
159 & \textit{6.00} &  \textit{-} & \textit{2.46} & \textit{1.00} & \textit{2.57} & \textit{1.00} & \textbf{\textit{2.18}} & \textit{1.00} \tabularnewline
160 & \textit{4.00} &  \textit{-} & \textbf{\textit{1.59}} & \textit{0.90} & \textit{2.39} & \textit{0.90} & \textit{1.75} & \textit{0.90} \tabularnewline
Mean & \textit{7.80} &  \textit{-} & \textit{2.00} & \textit{0.85} & \textit{2.55} & \textit{0.84} & \textbf{\textit{1.85}} & \textit{0.86} \tabularnewline
StdDev & \textit{5.62} &  \textit{-} & \textit{0.72} & \textit{0.20} & \textit{0.95} & \textit{0.20} & \textbf{\textit{0.67}} & \textit{0.19} \tabularnewline
Median & \textit{5.50} &  \textit{-} & \textit{1.83} & \textit{0.92} & \textit{2.34} & \textit{0.92} & \textbf{\textit{1.77}} & \textit{1.0} \tabularnewline
1st quartile & \textit{3.38} &  \textit{-} & \textit{1.52} & \textit{0.77} & \textit{1.96} & \textit{0.70} & \textbf{\textit{1.42}} & \textit{0.77} \tabularnewline
3rd quartile & \textit{13.63} &  \textit{-} & \textit{2.12} & \textit{1.00} & \textit{2.82} & \textit{1.00} & \textbf{\textit{1.98}} & \textit{1.00} \tabularnewline
\end{tabular}}
\end{table*}

\begin{table*}[!htb]
\centering
\caption{Quantitative results on the external validation set during the MICCAI 2022 edition (MAE - Median Absolute Error). The initial errors are reported in Table~\ref{tab:results}.}
\renewcommand{\arraystretch}{1.0}
\footnotesize
\resizebox{0.99\textwidth}{!}{%
\begin{tabular}{ccccccccc}
\label{tab:results_miccai}
Case & \multicolumn{2}{c}{Affine} &
\multicolumn{2}{c}{LapIRN} & \multicolumn{2}{c}{LapIRN + IO} &
\multicolumn{2}{c}{LapIRN + IO + IC}
\tabularnewline
\multicolumn{1}{c}{ID} & \multicolumn{1}{c}{MAE} & \multicolumn{1}{c}{Robustness} & \multicolumn{1}{c}{MAE} & \multicolumn{1}{c}{Robustness} & \multicolumn{1}{c}{MAE} & \multicolumn{1}{c}{Robustness} & \multicolumn{1}{c}{MAE} & \multicolumn{1}{c}{Robustness}
\tabularnewline
\hline
141 &  \textit{4.39} & \textit{1.00} & \textit{2.62} & \textit{1.00} & \textit{1.82} & \textit{1.00}  & \textbf{\textit{1.82}} & \textit{1.00} \tabularnewline
142 & \textit{6.13} & \textit{0.88} & \textit{3.12} & \textit{1.00} & \textit{2.79} & \textit{1.00} & \textbf{\textit{1.92}} & \textit{1.00} \tabularnewline
143 & \textit{8.50} & \textit{0.88} & \textit{3.07} & \textit{1.00} & \textit{2.56} & \textit{1.00} & \textbf{\textit{1.88}} & \textit{1.00} \tabularnewline
144 & \textit{9.46} & \textit{0.88} & \textit{2.99} & \textit{1.00} & \textbf{\textit{2.31}} & \textit{1.00} & \textit{2.62} & \textit{1.00} \tabularnewline
145 & \textit{4.99} & \textit{1.00} & \textit{2.15} & \textit{1.00} & \textit{1.10} & \textit{1.00} & \textbf{\textit{1.08}} & \textit{1.00} \tabularnewline
146 & \textit{6.40} & \textit{1.00} & \textit{2.33} & \textit{1.00} & \textit{1.73} & \textit{1.00} & \textbf{\textit{1.66}} & \textit{1.00} \tabularnewline
147 & \textit{2.17} & \textit{0.15} & \textit{2.34} & \textit{0.35} & \textit{2.11} & \textit{0.45} & \textbf{\textit{1.64}} & \textit{0.45} \tabularnewline
148 & \textit{2.86} & \textit{0.80} & \textit{2.00} & \textit{0.85} & \textbf{\textit{1.57}} & \textit{0.80} & \textit{1.67} & \textit{0.90} \tabularnewline
149 & \textit{2.34} & \textit{1.00} & \textit{2.10} & \textit{1.00} & \textit{1.65} & \textit{1.00} & \textbf{\textit{1.61}} & \textit{1.00} \tabularnewline
150 & \textit{3.99} & \textit{0.37} & \textit{2.78} & \textit{0.47} & \textit{1.34} & \textit{0.63} & \textbf{\textit{1.28}} & \textit{0.63} \tabularnewline
151 & \textit{2.10} & \textit{0.60} & \textit{1.71} & \textit{0.60} & \textbf{\textit{1.30}} & \textit{0.85} & \textit{1.31} & \textit{0.85} \tabularnewline
152 & \textit{2.14} & \textit{0.95} & \textit{1.65} & \textit{0.84} & \textbf{\textit{1.43}} & \textit{0.89} & \textit{1.51} & \textit{0.89} \tabularnewline
153 & \textit{1.90} & \textit{0.50} & \textit{2.00} & \textit{0.42} & \textit{1.74} & \textit{0.67} & \textbf{\textit{1.66}} & \textit{0.75} \tabularnewline
154 & \textit{2.45} & \textit{0.30} & \textit{1.99} & \textit{0.45} & \textit{2.03} & \textit{0.40} & \textbf{\textit{1.98}} & \textit{0.45} \tabularnewline
155 & \textit{2.88} & \textit{0.22} & \textit{2.91} & \textit{0.42} & \textbf{\textit{2.01}} & \textit{0.42} & \textit{2.03} & \textit{0.42} \tabularnewline
156 & \textit{3.29} & \textit{1.00} & \textit{2.79} & \textit{1.00} & \textbf{\textit{1.48}} & \textit{1.00} & \textit{1.53} & \textit{1.00} \tabularnewline
157 & \textit{5.86} & \textit{0.90} & \textit{2.99} & \textit{1.00} & \textbf{\textit{1.52}} & \textit{1.00} & \textit{1.58} & \textit{1.00} \tabularnewline
158 & \textit{3.75} & \textit{0.40} & \textit{2.56} & \textit{0.80} & \textit{1.16} & \textit{1.00} & \textbf{\textit{1.16}} & \textit{1.00} \tabularnewline
159 & \textit{7.78} & \textit{0.36} & \textit{2.75} & \textit{1.00} & \textbf{\textit{2.30}} & \textit{1.00} & \textit{2.37} & \textit{1.00} \tabularnewline
160 & \textit{2.72} & \textit{0.80} & \textit{2.66} & \textit{0.70} & \textbf{\textit{1.79}} & \textit{0.90} & \textit{1.80} & \textit{0.90} \tabularnewline
Mean & \textit{4.31} & \textit{0.70} & \textit{2.48} & \textit{0.70} & \textit{1.79} & \textit{0.85} & \textbf{\textit{1.71}} & \textit{0.86} \tabularnewline
StdDev & \textit{2.32} & \textit{0.30} & \textit{0.46} & \textit{0.25} & \textit{0.46} & \textit{0.22} & \textbf{\textit{0.38}} & \textit{0.21} \tabularnewline
Median & \textit{3.52} & \textit{0.84} & \textit{2.59} & \textit{0.93} & \textit{1.74} & \textit{1.00} & \textbf{\textit{1.66}} & \textit{1.00} \tabularnewline
1st quartile & \textit{2.42} & \textit{0.39} & \textit{2.08} & \textit{0.57} & \textbf{\textit{1.47}} & \textit{0.77} & \textit{1.53} & \textit{0.83} \tabularnewline
3rd quartile & \textit{5.93} & \textit{0.96} & \textit{2.82} & \textit{1.00} & \textit{2.05} & \textit{1.00} & \textbf{\textit{1.89}} & \textit{1.00} \tabularnewline

\end{tabular}}
\end{table*}

\section{Discussion and Conclusion}
\label{sec:disc}
The results confirm that the proposed hybrid approach, based on both deep learning and instance optimization (IO), improves the quantitative results compared to the methods used separately. Interestingly, the IO alone (extended by the IO-based affine registration) provides slightly better results than the LapIRN without any further tuning. This is expected because the IO optimizes each case separately and we do not use any supervision other than the unsupervised objective function to guide the training process. Interestingly, the worst robustness is reported for cases with a low initial error, suggesting that the unsupervised registration cannot improve the results more.

The objective function weighting further improves the results. From the quantitative perspective the difference is not significant, however, it is connected with the characteristic of target registration error (TRE). The TRE is not dedicated to the evaluation of the image registration with missing data because the corresponding points cannot be annotated directly in the missing volume. Therefore, the correlation between the TRE and the qualitative results nearby the tumor is limited.

Our method has several limitations. First, we do not use information about sparse landmarks during training. We think that it would be possible to use the landmarks as a weak-supervision, however, with MAE at the level of 1.71 pixel, there is not much space for further improvements, and such a method would probably learn only the annotators' preferences. Second, we did not extensively explored the use of different modalities. Probably the use of modality used for the annotation would slightly improve the quantitative results. What is more, tuning the hyperparameters to just a subset of the available modalities could possibly further improve the results. Moreover, the inverse consistency weighting significantly increases the registration time (from several seconds to almost one minute). However, we think that future work could introduce the objective function weighting directly into the deep network training and significantly decrease the registration time.

In our opinion, several decisions regarding the challenge setup should be rethought. First, the MAE and the related robustness should not be used as the only evaluation criteria for image registration with missing data. The evaluation landmarks cannot be chosen directly within the tumor volume because it is missing in the target scan. Additional metrics, based e.g. on the relative tumor volume~\cite{btb} could be used to evaluate the registration quality. Second, it could be considered to score the submissions using the displacement field smoothness outside the tumor volume (and the recurrences), as well as the registration time. Moreover, uploading the displacement fields directly instead of the transformed landmarks would enable the organizers to calculate all the statistics automatically using the challenge evaluation platform.

To conclude, we proposed a hybrid approach based on the DL and IO as a contribution to the BraTS-Reg challenge. We have shown that the proposed method improves the alignment for majority of the external validation cases with high robustness. The proposed objective function weighting based on the inverse consistency further improved the results. The method could be further extended by directly incorporating the unsupervised objective function weighting during the deep network training.

\subsubsection{Acknowledgements and Compliance With Ethical Standards}
\label{sec:acknowledgments}

This research study was conducted retrospectively using human subject data made available in open access by BraTS-Reg Challenge organizers~\cite{dataset}. Ethical approval was not required as confirmed by the license attached with the open access data. The authors declare no conflict of interest. The authors would like to thank the organizers for their help with the submission of the Singularity container. This research was supported in part by PLGrid Infrastructure.

\bibliographystyle{splncs04}
\bibliography{main}

\begin{thebibliography}{10}
\providecommand{\url}[1]{\texttt{#1}}
\providecommand{\urlprefix}{URL }
\providecommand{\doi}[1]{https://doi.org/#1}

\bibitem{syn}
Avants, B., et~al.: {Symmetric diffeomorphic image registration with
  cross-correlation: evaluating automated labeling of elderly and
  neurodegenerative brain}. Medical Image Analysis  \textbf{12},  26--41 (2008)

\bibitem{dataset}
Baheti, B., Waldmannstetter, D., Chakrabarty, S., et~al.: {The Brain Tumor
  Sequence Registration Challenge: Establishing Correspondence between
  Pre-Operative and Follow-up MRI scans of diffuse glioma patients}. {arXiv}
  (2112.0697) (2021)

\bibitem{u2}
Balakrishnan, G., et~al.: {VoxelMorph: A Learning Framework for Deformable
  Medical Image Registration}. IEEE Transactions on Medical Imaging
  \textbf{38}(8),  1788--1800 (2019)

\bibitem{u_voxemorph_diffeo_2}
Dalca, A., et~al.: {Unsupervised learning of probabilistic diffeomorphic
  registration for images and surfaces}. Medical Image Analysis  \textbf{57},
  226--236 (2019)

\bibitem{pylightning}
Falcon, W., Cho, K.: {A Framework For Contrastive Self-Supervised Learning And
  Designing A New Approach}. arXiv preprint arXiv:2009.00104  (2020)

\bibitem{hering2021learn2reg}
Hering, A., et~al.: Learn2reg: comprehensive multi-task medical image
  registration challenge, dataset and evaluation in the era of deep learning.
  arXiv preprint arXiv:2112.04489  (2021)

\bibitem{synthmorph}
Hoffmann, M., et~al.: {SynthMorph: Learning Contrast-Invariant Registration
  Without Acquired Images}. IEEE Transactions on Medical Imaging  (2021),
  {Early Access}

\bibitem{mok}
Mok, T., Chung, A.: {Fast Symmetric Diffeomorphic Image Registration with
  Convolutional Neural Networks}. {IEEE CVPR} pp. 4644--4653 (2020)

\bibitem{mok2}
Mok, T., Chung, A.: {Large Deformation Diffeomorphic Image Registration with
  Laplacian Pyramid Networks}. MICCAI 2020 pp. 1--10 (2020)

\bibitem{source_code}
Wodzinski, M., Jurgas, A.: {Source Code}.
  \url{https://github.com/MWod/BraTS_Reg} (2022)

\bibitem{btb}
Wodzinski, M., et~al.: {Semi-Supervised Deep Learning-based Image Registration
  Method with Volume Penalty for Real-time Breast Tumor Bed Localization}.
  Sensors  \textbf{21}(12),  1--14 (2021)

\end{thebibliography}

\end{document}